\documentclass{article} % For LaTeX2e
\usepackage{bbm}
\usepackage{graphicx}
\usepackage{subfigure}
\usepackage{hyperref}
\usepackage{multirow}
\usepackage{xcolor}
\usepackage{xspace}
\usepackage{makecell}
\usepackage{amssymb}
\usepackage{url}

\newcommand{\dataset}[1]{\textit{#1}\xspace}
\newcommand{\best}[1]{\textbf{#1}}

\newcommand{\method}{\textit{SANP}\xspace}
\usepackage{iclr2020_conference,times}
\usepackage{amsmath,amsfonts,bm}

\def\eqref#1{equation~\ref{#1}}
\def\1{\bm{1}}

\DeclareMathAlphabet{\mathsfit}{\encodingdefault}{\sfdefault}{m}{sl}
\SetMathAlphabet{\mathsfit}{bold}{\encodingdefault}{\sfdefault}{bx}{n}

\usepackage{hyperref}
\usepackage{url}
\usepackage{graphicx}
\usepackage{subfigure}

\title{Self-Adaptive Network Pruning}

\author{
    Jinting Chen \textsuperscript{1}
    , Zhaocheng Zhu \textsuperscript{2}
    , Cheng Li \textsuperscript{1}
    , Yuming Zhao \textsuperscript{1}
    \thanks{Corresponding author.} \\
Department of Automation, Shanghai Jiao Tong University \textsuperscript{1} \\
Mila - Qu\'ebec AI Institute \textsuperscript{2}
}

\begin{document}

\maketitle

\begin{abstract}
Deep convolutional neural networks have been proved successful on a wide range of tasks, yet they are still hindered by their large computation cost in many industrial scenarios. In this paper, we propose to reduce such cost for CNNs through a self-adaptive network pruning method (\method). Our method introduces a general Saliency-and-Pruning Module (SPM) for each convolutional layer, which learns to predict saliency scores and applies pruning for each channel. Given a total computation budget, \method adaptively determines the pruning strategy with respect to each layer and each sample, such that the average computation cost meets the budget. This design allows \method to be more efficient in computation, as well as more robust to datasets and backbones. Extensive experiments on 2 datasets and 3 backbones show that \method surpasses state-of-the-art methods in both classification accuracy and pruning rate.
\end{abstract}

\section{Introduction}

Recently, convolutional neural networks (CNNs) have become a dominant approach in a wide range of visual tasks. Typical applications of CNNs include image classification~\cite{krizhevsky2012imagenet}, object detection~\cite{girshick2014rich} and semantic segmentation~\cite{long2015fully}. Despite their success, it is still a challenge to deploy CNNs in industrial scenarios. This is mainly because CNNs are designed to be over-parameterized~\cite{du2018gradient}, and require much computation during inference. For example, ResNet-18, the smallest version of ResNet~\cite{he2016deep}, requires 2 GFLOPs for a single prediction, which is unaffordable for most smartphones or embedded systems.

To reduce the computation demand of CNNs, many methods have been proposed from several perspectives. A bunch of methods~\cite{chollet2017xception}\cite{zhang2018shufflenet}\cite{ma2018shufflenet} propose to build efficient architectures with depthwise separable convolutions. Some methods~\cite{courbariaux2016binarized}\cite{zhou2016dorefa}\cite{micikevicius2017mixed} learn models for low-precision inference. However, these methods require careful design of models or quantization functions, which can hardly generalize to other tasks without heavy engineering. Most recently, there are a number of methods \cite{han2015deep}\cite{li2016pruning}\cite{NS} that try to prune the connections in networks. These methods drop parameters or channels according to some saliency scores, such as $L_1$-norm values of parameters or channels. As the scores are adaptively computed with regard to the model as well as the task, these methods can be easily applied to different scenarios. Therefore, we also follow this stream in this paper, and propose a novel self-adaptive pruning method.

\begin{figure}[!h]
  \centering 
  \subfigure{ 
    \label{fig:NS:a}
    \includegraphics[height=3.5cm,width=6cm]{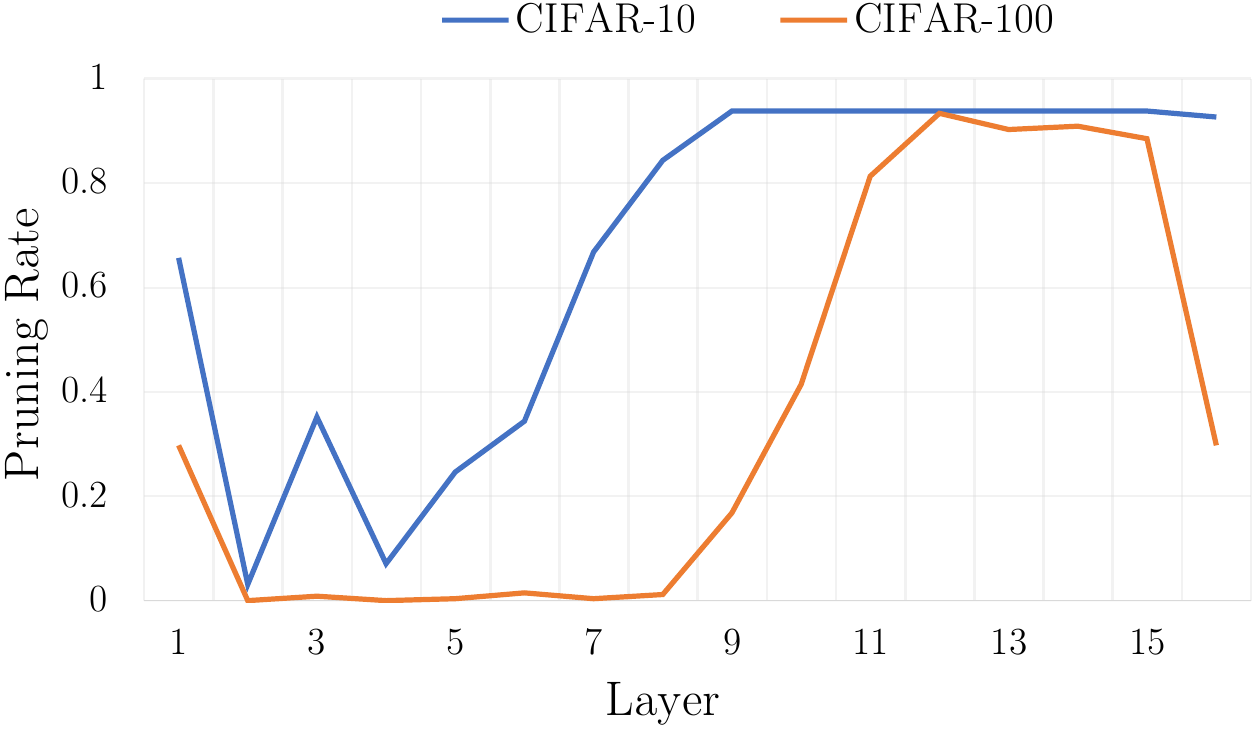}
  }
  \subfigure{ 
    \label{fig:NS:b}
    \includegraphics[height=3.5cm,width=6cm]{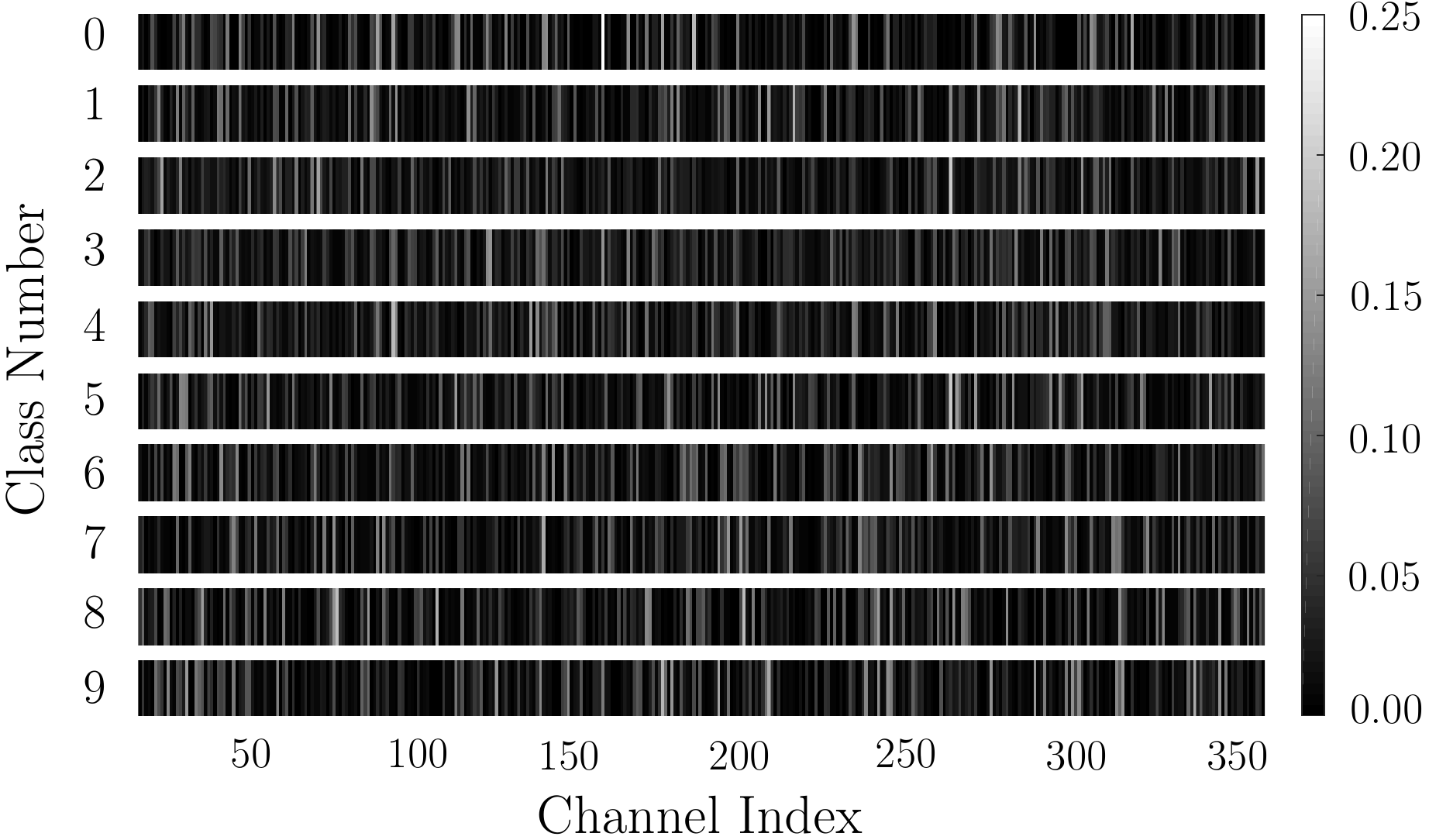} 
  }
  \caption{
  Visualization of channel activations in a pruned VGGNet.
  (a) Pruning rate w.r.t. layers. The pruning rate varies from layer to layer on both datasets.
  (b) Average activation of each channel in the last convolutional layer w.r.t. categories in \dataset{CIFAR-100}. Only a small subset of channels is activated for each category.}
  %The average Pearson correlation coefficient between all category pairs is \textcolor{red}{xxxxx}.
  %\caption{(a)The pruning ratio per layer of VGGNet on \dataset{CIFAR-10} and \dataset{CIFAR-100} respectively after pruned by NS\cite{NS}. (b)The averaged activation scores of channels in the last convolutional layer with regard to each category in \dataset{CIFAR-100}. This layer contains 360 channels after pruned by NS\cite{NS}} 
  \label{fig:NS}
\end{figure}

Typically, a network pruning recipe at channel level consists of 2 ingredients, a saliency estimation module and a pruning module. Given a budget, the pruning algorithms \cite{li2016pruning}\cite{NS} first learn saliency scores for each channel, and then prune channels that have low scores. However, we argue this formulation is not enough for a good pruning. Take VGGNet \cite{simonyan2014very} pruned by Network Slimming (NS) \cite{NS} as an example, we observe two phenomena:

\begin{enumerate}
    \item{For different layers, the optimal pruning rates are very different.}
    \item{For each category, it only activates a small subset of remaining channels.}
\end{enumerate}

Figure \ref{fig:NS} illustrates these phenomena. The first phenomenon shows that there does not exist a constant pruning rate for every layer. In other words, layers would be either over-pruned or under-pruned by any global pruning rate. This is because most CNN architectures are designed for \dataset{ImageNet}, and the capacity of layers does not necessarily fit \dataset{CIFAR-10}, \dataset{CIFAR-100} and other datasets. Hence, a good pruning strategy should set different pruning rates for each layer. The second phenomenon indicates that a static pruning strategy is sub-optimal, since only a small set of channels is required for each category. To get better pruning performance, the pruning strategy needs to be conditioned on the input images. Ideally, we would like to have a pruning method that has both properties.

In this paper, we propose a self-adaptive method (\method) for network pruning. Our method satisfies the above two properties through a layer-adaptive and sample-adaptive design. Specifically, the layer adaptiveness is achieved by a cost estimation step for each layer, with only budget constraints on the total computation cost. The sample adaptiveness is achieved by a saliency prediction step over the current input sample. Both steps utilize differentiable modules and thereby can be jointly trained with classification objective using a multi-task loss. Our method adaptively determines the computation routine for each layer and each sample, and improves the pruning rate over state-of-the-art methods, without sacrifice on performance. The contribution of this paper is three folds:

\begin{enumerate}
    \item {We propose a novel method \method for network pruning, which adaptively learns the pruning rate for each layer and each sample.}
    \item {We instantiate \method with differentiable modules, and enable joint training with classification and cost objectives.}
    \item {We empirically evaluate our method on 2 datasets and 3 backbones and it achieves state-of-the-art performance in all settings.}
\end{enumerate}

\section{Related Works}

%\textcolor{red}{Zhu: We'd better put a brief summary sentence here, and then walk into small domains. For methods that satisfy one of the adaptiveness, please point it out. Also point out that it doesn't satisfy the other.}
%Existing network pruning methods can be categorized into two groups: static network pruning and dynamic path network.

\subsection{Static Network Pruning}

%Previous
Static pruning methods generate a fixed network for all novel images. They can be divided into weight pruning methods and channel pruning methods. Weight pruning methods work on pruning fine-grained weights of the filters, resulting in unstructured sparsity. For example, Han et al.~\cite{han2015deep} iteratively prune near-zero weights to obtain a pruned network without loss of precision. Channel pruning methods reduce model size at channel level and can achieve a sparse structure. \cite{li2016pruning} iteratively prunes filters whose $L_1$-norm values are relatively small and retrains the remaining network. NS \cite{NS} introduces sparsity on the scaling parameters of Batch Normalization (BN) layers and proposes an iterative two-step algorithm to prune the network. AutoPruner~\cite{AutoPruner} integrates channel pruning and model fine-tuning into a single end-to-end trainable framework. Filter Clustering and Pruning (FCP)~\cite{OFCP} adds an extra cluster loss to the loss function, which forces the filters in each cluster to be similar and thereby prunes redundant channels. NS, AutoPruner and FCP could adaptively determine pruning rate for each layer, but their pruning strategies are invariant with regard to different samples.

\subsection{Dynamic Path Network}

Instead of using the entire feed forward graph of the network, dynamic path networks~\cite{lin2017runtime}\cite{liu2018dynamic}\cite{CG}\cite{FBS} selectively execute a subset of modules at inference time based on input samples. Runtime Neural Pruning~\cite{lin2017runtime} uses an agent to judge channel importance and prunes unimportant channels according to different samples with reinforcement learning. Liu et al.~\cite{liu2018dynamic} propose a dynamic deep neural network to execute a subset of neurons and use deep Q-learning to train the controller modules. The above dynamic networks train their strategies through reinforcement learning because the binary decisions cannot be represented by differentiable functions. Therefore these methods are hard to generalize on multiple datasets and networks. Recently, several methods overcome this limitation. Channel Gating (CG)~\cite{CG} splits channels in each layer into two groups and the proportion of the first group is uniform for all layers. Then it identifies ineffectual receptive fields based on the first group of channels and skips computation on the second group in these fields. It uses continuous functions to approximate the gradient of non-differentiable binary functions. Feature Boosting and Suppression (FBS) \cite{FBS} sets a constant pruning rate for each layer and amplifies salient channels based on the current input sample. It utilizes a k-winners-take-all function which is partially differentiable. Though both CG and FBS are sample-adaptive methods, they lack layer-adaptiveness to regulate pruning rate for different layers.  
\section{Our Method}
\begin{figure}
    \centering
    %\subfigure[Overall pipeline]{
    %\includegraphics[height=2.5cm]{figure/pipeline-6.pdf}
    %}
    %\subfigure[Layer pipeline]{
    %\includegraphics[height=3cm]{figure/SPM-6.pdf}
    %}
    \includegraphics[height=6.5cm]{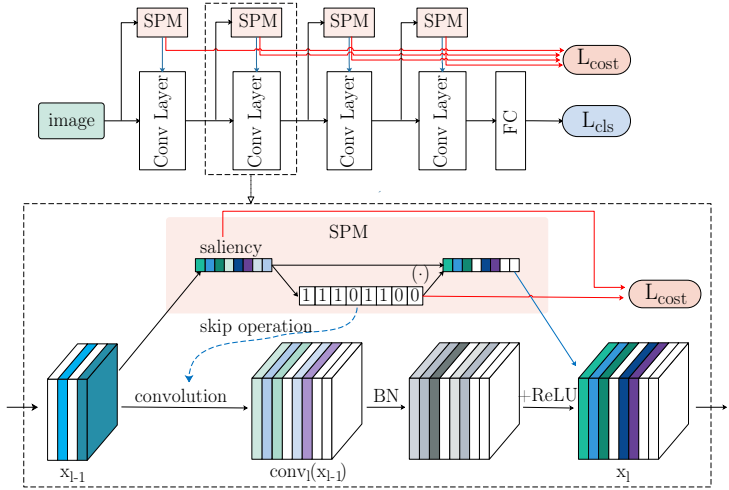}
    \caption{The overall pipeline and layer pipeline of \method. Colors of channels indicate their saliency scores, where white denotes zero saliency. First, the input features are down-sampled and passed into SPM to get saliency scores and pruning decisions for channels. Then channel pruning decisions are used to estimate current computation cost. The estimation automatically adjusts importance of classification loss and cost loss so that the network adaptively determines pruning rate for each layer with regard to the computation budget.}
    %\textcolor{red}{Zhu: Briefly introduce the computation flow in two sentences here. e.g. First, the input layer is downsampled and passed into SPM module to get ...}}
    %\caption{we use the $l^{th}$ layer with 4-channel input and 8-channel output, where channels are colored to indicate saliency scores, and the white blocks represent all-zero channels. $(\cdot)$ here denotes element-wise multiplication.}
    \label{fig:pipeline}
\end{figure}

Figure \ref{fig:pipeline} shows the main pipeline of \method. Firstly, Saliency-and-Pruning Module is embedded in each convolutional layer of backbone network. It predicts saliency scores for channels based on input features and then generates pruning decision for each channel. The convolution operation would be skipped for these channels whose corresponding pruning decision is 0, as indicated by the dashed arrow. 
Then we jointly train the backbone network and SPMs with both classification objective and cost objective. We estimate computation cost dependent on the pruning decisions in each layer. The estimation adjusts importance of two objectives so that network could adaptively determine pruning rate per layer with a total computation budget.
Since input features and output features are both sparse, the expensive convolution operation can be accelerated from both sides. 
Then we will go into details about the proposed method.

\subsection{Saliency-and-Pruning Module}
To prune channels layer-by-layer, some modules or separate networks are required. And the pruning strategy must be decided before each layer is activated. We propose a Saliency-and-Pruning Module, a lightweight network module for this purpose. 

Since we observe that only a subset of channels is actually activated for different categories, we decide to determine channel pruning strategy dependent on each input image to get better performance. And in order to adaptively find the most important channels for each sample, we generate saliency scores for kernels based on the input features from previous layer $x^{l-1} \in \mathbb{R}^ {C_{l-1} \times H_{l-1}\times W_{l-1}}$. Saliency prediction can be defined as
\begin{equation}
s^l(x^{l-1}) = SaliencyPrediction(x^{l-1},W)
\end{equation}
where $SaliencyPrediction(\cdot)$ denotes saliency function. In Section \ref{sec:saliency}, we would introduce this function specifically. 

Then the channels with low saliency scores could be pruned so as to accelerate backbone network. We need to adopt 0/1 binary valued function to decide whether the calculation of each channel is skipped or not. However, binary functions are not differentiable and thereby these problems are usually approached with reinforcement learning. Unlike previous work, we utilize a discretization technique called Improved Semantic Hashing \cite{kaiser2018discrete}, which enables classification loss back-propagate to SPMs. Hence, the backbone network and SPMs could be jointly trained in an end-to-end manner. The pruning decisions could be formulated as: 
\begin{equation}
 b^l(x^{l-1})=Binarize(s^{l-1}) 
\end{equation}
where $Binarize(\cdot$) denotes binarization function.In Section \ref{sec:binarization}, we would introduce this function specifically. 

Further, channels with higher saliency scores are naturally the more significant channels, therefore we propose to rescale the output features with saliency scores to make these channels more decisive. 
% We leveraging the saliency after BN layers.
Since modern deep neural networks~\cite{he2016deep}\cite{huang2017densely}\cite{zhang2018shufflenet} apply BN layers after convolutional layers, we leverage the rescaling operation directly after each BN layer. Here, we propose to define the calculation of a batch-normalized convolutional layer with SPM. The $i^{th}$ channel of output features  $x^{l}\in\mathbb{R}^{C_l\times H_l\times W_l}$ is formulated as:
\begin{equation}
 x_i^l=s_i^l(x^{l-1})\cdot b_i^l(x^{l-1})\cdot BatchNorm(f_i^l\ast x^{l-1}) 
\end{equation}
where $f_i^l$ denotes the $i^{th}$ convolutional kernel of $l^{th}$ layer and $\ast$ denotes convolution operation. Since $b^l$ is a binary code, we can reformulate the computation of $x_i^l$:
\begin{equation} x_i^l=
\left\{
             \begin{array}{lr}
             \textbf{0}, &  if\ \ b_i^l=0  \\
              s_i^l(x^{l-1})\cdot b_i^l(x^{l-1})\cdot BatchNorm(f_i^l\ast x^{l-1}) , & if  \ \ b_i^l\neq0
             \end{array}
\right.
\label{eq:convolution}
\end{equation}
Here \textbf{0} is a 2-D feature map with all its elements being 0. That is, if  $b_i^l$=0, convolution operation of filter $f_i^l$ is skipped and \textbf{0} is used as the output instead. All convolutions can take advantage of both input-side and output-side sparsity. Additionally, $C_l$ denotes the channel number of $x^l$, $H_l$  and $W_l$ denotes the height and width of feature map.

\subsection{Saliency Function}
\label{sec:saliency}
To obtain saliency, we firstly use global average pooling to squeeze global spatial information into a channel descriptor following SEblock \cite{SENet}. Specifically, the channel descriptor $d\in\mathbb{R}^{C_{l-1}}$ is calculated by the following formula:
\begin{equation}
d=\frac{1}{H_{l-1}\times W_{l-1}}\sum_{i=1}^{H_{l-1}}\sum_{j=1}^{W_{l-1}}x^{l-1}(i,j)
\end{equation}
Then we consider using fully-connected layers to map channel-wise statics to predict saliency scores for kernels in $l^{th}$ layer. In order to reduce computation, we use a reduction rate r like SEblock\cite{SENet}. The saliency scores $s\in\mathbb{R}^{C_l}$ can be defined as:
\begin{equation}
 s^l(x^{l-1})=SaliencyPrediction(x^{l-1},W)=W_2\delta(W_1d) 
 \end{equation}
where $\delta$ refers to the ReLU function, $W_1 \in \mathbb{R}^{\frac{C_l}{r}\times C_{l-1}}$, $W_2\in\mathbb{R}^{C_l\times\frac{C_l}{r}}$.
%The results in Section 4 use the r=4 by default.

\subsection{Binarization Function}
\label{sec:binarization}
We adopt a recently proposed discretization technique namely Improved Semantic Hashing \cite{kaiser2018discrete} to generate channel pruning strategy from saliency scores $s^l$($x^{l-1}$). 

During training, a Gaussian noise $\xi \sim N(0,1)^{C_l}$ is add to $s^l(x^{l-1})$. As with all of the operations below, the sum operation is element-wise. 
Then we compute the vector through a saturating sigmoid function \cite{kaiser2015neural}:
\begin{equation}
 s_1 = \max(0, \min(1, a \cdot \sigma(s^l(x^{l-1}) + \xi) - b))
\end{equation}
where $\sigma$ denotes the original sigmoid function, $a$ and $b$ denote hyperparameters.

The binary code is then constructed via rounding:
\begin{equation}
s_2=\mathbbm{1}(s_1 > 0.5)
\end{equation}
In the forward propagation, we use $s_1$ half of the time and $s_2$ the other half. $s_2$ is computed from non-differentiable function. Therefore in the backward propagation, we let gradients always flow to $s_1$, even if $s_2$ is used in the forward propagation. 

During evaluation and inference, $s_2$ is used all the time. Note that the Gaussian noise is only used for training and we set $\xi$ to \textbf{0} during evaluation and inference.

\subsection{Multi-task Training}
Until now, one question remains that how to control sparsity of the network, so that it reaches a computation budget. We observe that the optimal pruning rate varies with different layers and thereby the pruning method should adaptively learn pruning rate for each layer. To solve this problem, we 
propose a multi-task training with both classification objective and cost objective. We induce network sparsity with $L_1$-norm on saliency scores and estimate the current computation cost with pruning decisions generated by SPMs in each layer. The cost estimation adjusts importance of two objectives so that network could adaptively determine pruning rate for each layer with a total computation budget. The multi-task loss could be formulated as:
\begin{equation}   \mathcal{L}_{multi}=\mathcal{L}_{cls}+\lambda\frac{1}{N_c}\sum_{l=1}^L\Vert s^l\Vert_1 
\end{equation}
where the first term is a classification loss (e.g., cross entropy loss), and the second term is the cost loss $\mathcal{L}_{cost}$. $N_c$ denotes total filter number of backbone network, $L$ denotes total layers of backbone network.

The value of $\lambda$ is automatically adjusted according to the estimation of current computation cost:
\begin{equation}
\lambda = \lambda_0 \cdot \frac{(p_t-p)}{p_0}
\label{eq:lambda}
\end{equation}
where $p_t$ is the estimation of current computation cost, calculated from binary code $b^l$ of each layer. In practice, we collect several estimated values during training, and then calculate $p_t$ based on these data. $p$ is the given budget, $p_0$ is computation cost of the total network. $\lambda_0$ is a constant, and the range of $\lambda$ is $[-\lambda_0,\lambda_0]$ according to equation \ref{eq:lambda}. 

If current computation cost is far from expectation, then $\lambda$ is relatively large and thus the training could pay more attention to cost loss. In more detail, if $p_t>p$, then $\lambda$ is positive and the network becomes more sparse so that $p_t$ would decline. Otherwise, $\lambda$ is negative and the network becomes less sparse so that $p_t$ would increase. Once $p_t$ is rather close to the budget, $\lambda$ is relatively low, which means the network can focus on classification task. The actual obtained computation cost can be close to the budget, but not necessarily equal to it.

\section{Experiments}
\subsection{Experiment Setup}
\subsubsection{Datasets and evaluation metrics}
We evaluate our method on \dataset{CIFAR-10} and \dataset{CIFAR-100}~\cite{CIFAR}. Both datasets contain 60,000 32$\times$32 colored images, with 50,000 images for training and 10,000 for testing. They are labeled for 10 and 100 classes in \dataset{CIFAR-10} and \dataset{CIFAR-100} respectively.

Classification performance is measured by top-1 accuracy and computation cost is evaluated by the floating-point operations (FLOPs). The FLOPs  of $l^{th}$ convolutional layer in inference is calculated as $FLOPs=HW(C_{in}k^2+1)C_{out}$. $H$, $W$, $C_{out}$ is the height, width and channel number of output features, $k$ is the kernel size, $C_{in}$ is channel number of input features and 1 refers to bias. 

\subsubsection{Implementation details}
We use M-CifarNet \cite{zhao2018mayo}, VGGNet \cite{simonyan2014very}, ResNet-18 \cite{he2016deep} as backbone networks in our experiments. The additional computation required for SPMs in inference is approximately 0.01\% of the total network. We adopt PyTorch for implementation and utilize Momentum SGD as the optimizer. We use a batch size of 256 for \dataset{CIFAR-10} and \dataset{CIFAR-100}. We set the initial learning rate to 0.1 and decrease it by a factor of 10 every 100 epochs. $\lambda_0$ is set to 0.01 in our experiments. The backbone network is firstly trained to match state-of-the-art performance on those datasets. Then we replace all batch-normalized convolutional layer calculations with equation \ref{eq:convolution} and initialize the convolution kernels with the pre-trained weights.
Afterwards, we warm up the SPMs using the classification loss, with fixed parameters of convolutional kernels.
Finally, we jointly fine-tune backbone and SPMs with multi-task loss to meet the computation budget, as well as maximize the accuracy.
%Afterwards, we freeze parameters of convolutional kernels and train SPMs with classification loss only. 
%This is to minimize the influence of pre-trained backbone network due to premature SPMs.
%Finally, we jointly fine-tune backbone and SPMs with multi-task loss to preserve accuracy as much as possible and achieve the predefined computational cost.

\subsection{Experiment Results}
%\subsubsection{\method vs. state-of-the-art methods}

\begin{table}[!h]
    \centering
    \begin{tabular}{|c|c|c|c|c|c|c|}
        \hline
        Backbone & Model & L-a & S-a & Error(\%) & FLOPs(M) & Pruned(\%) \\
        \hline
        \multirow{3}{*}{M-CifarNet} & Unpruned & & & 8.63 & 174.3 & \\
         & \cite{FBS} & & \checkmark & \best{9.41} & 44.3 & 74.6 \\
        & \method & \checkmark & \checkmark & \best{9.41} & \best{39.3} & \best{77.5} \\
        \hline
        \multirow{4}{*}{VGGNet} & Unpruned & & & 6.34 & 398.5 & \\
        & \cite{NS} & \checkmark & & 6.20 & 195.5 & 51.0 \\
        & \cite{OFCP} & \checkmark & & 6.24 & 143.9 & 63.9 \\
        & \method & \checkmark & \checkmark & \best{6.18} & \best{133.9} & \best{66.4} \\
        \hline
        \multirow{3}{*}{ResNet-18} & Unpruned & & & 5.40 & 501 & \\
        & \cite{CG} &  & \checkmark & \best{5.62} & 172 & 65.6 \\
        & \method & \checkmark & \checkmark & 5.64 & \best{163} & \best{67.5} \\
        \hline
    \end{tabular}
    \caption{Comparison of different methods on \dataset{CIFAR-10}. The best results from pruning methods are emphasized.
    %The FLOPs corresponds to the number of floating point computation in inference for a single image.
     `L-a' and `S-a' denote layer adaptiveness and sample adaptiveness respectively.}
    \label{tab:main_cifar10}
\end{table}

%\vspace{-2em}

\begin{table}[!h]
    \centering
    \begin{tabular}{|c|c|c|c|c|c|c|}
        \hline
        Backbone & Model &  L-a  &  S-a  & Error(\%) & FLOPs(M) & Pruned(\%) \\
        \hline
        \multirow{4}{*}{VGGNet} & Unpruned & & & 26.74 & 398.5 & \\ 
        & \cite{NS} & \checkmark & & 26.52 & 250.5 & 37.1 \\
        & \cite{OFCP} & \checkmark & & \best{26.45} & 196.3 & 50.7 \\  
        & \method & \checkmark & \checkmark & 26.47 & \best{170.6} & \best{57.2} \\
        \hline
        \multirow{3}{*}{ResNet-18} & Unpruned & & & 24.95 & 501 & \\
        & \cite{CG} &  & \checkmark & 25.24 & 200 & 59.9 \\
        & \method & \checkmark & \checkmark & \best{25.20} & \best{189} & \best{62.3} \\
        \hline
    \end{tabular}
    \caption{Comparison of different methods on \dataset{CIFAR-100}. The best results from pruning methods are emphasized. `L-a' and `S-a' denote layer adaptiveness and sample adaptiveness respectively. Note that M-CifarNet does not have unpruned baseline on \dataset{CIFAR-100}, and is ignored in this table.}
    %Note that M-CifarNet has not been utilized for pruning on \dataset{CIFAR-100} in previous work.
    \label{tab:main_cifar100}
\end{table}

We compare our method with several state-of-the-art network pruning methods on \dataset{CIFAR-10} and \dataset{CIFAR-100}: (1)FBS~\cite{FBS}, (2)NS~\cite{NS}, (3)FCP~\cite{OFCP}, and (4)CG~\cite{CG}. NS and FCP are layer-adaptive pruning methods while FBS and CG are sample-adaptive pruning methods. %CG skip calculation of insignificant receptive fields of input image, which is complementary to our method and can be utilized together with ours to speed up CNNs. 
Table \ref{tab:main_cifar10} and \ref{tab:main_cifar100} present results on \dataset{CIFAR-10} and \dataset{CIFAR-100} respectively. We observe that our method would perform better compared with these methods. With almost the same accuracy, our model uses less computational cost.

\section{Ablation Study}

In this section, we design several ablation studies to give a comprehensive understanding of the two adaptiveness in \method. For convenience, all experiments in this section are conducted on \dataset{CIFAR-10} and ResNet-18.

%\subsection{Essentiality of Adaptiveness}
\subsection{Are layer-adaptiveness and sample-adaptiveness necessary?}
%To analyze the impact of adaptiveness of \method, we compare \method with the following two variants.
In \method, we propose to prune channels adaptively for each layer and each input sample. To see whether such a design is necessary, we compare our method with two variants.

\begin{enumerate}
    \item FIXED K. This variant removes the binarization function of \method and always selects fixed $k$ percentage of channels with highest saliency scores for each layer. $k$ is predefined and we choose $k$ to match the expected computational cost. Therefore, the rate of activated channels is invariant across layers.
    \item STATIC. This variant generates saliency scores from the same static vector and thereby the channel pruning strategy is invariant for all input samples.
\end{enumerate}

%The results in Table \ref{tab:adaptiveness} indicates that both layer-adaptiveness and sample-adaptiveness are necessary for a good pruning model.
%\textcolor{red}{Check if correct.}
Table \ref{tab:adaptiveness} shows the results of \method and its variants. It is observed that under similar computation budget, \method achieves the best performance among all methods, indicating that both layer-adaptiveness and sample-adaptiveness are necessary for good pruning.

%Here we utilized ResNet-18 as backbone network and set $k$ to 0.8125.

\begin{table}
    \centering
        \begin{tabular}{|c|c|c|c|c|c|}
        \hline  
        Model & L-a & S-a & Test error(\%)  & FLOPs(M/image) & Pruned rate(\%) \\  
        \hline  
        FIXED K & & \checkmark & 5.62 & 331 & 33.9 \\
        STATIC & \checkmark & & 5.48 & 337 & 32.8 \\
        \method & \checkmark & \checkmark & \best{5.41} & 333 & 33.5 \\
        \hline
        \end{tabular}
    \caption{Results of test error and computation cost by different designs. \method with layer-adaptiveness and sample-adaptiveness gets the best performance.}
    %\caption{Impact of adaptiveness on \dataset{CIFAR-10} with ResNet-18 as backbone network.}
    \label{tab:adaptiveness}
\end{table}  

\subsection{What is the distribution of pruned channels?}

\begin{figure}[!h]
    \begin{minipage}[t]{0.47\linewidth}
    \centering
    \includegraphics[height=3.3cm,width=6cm]{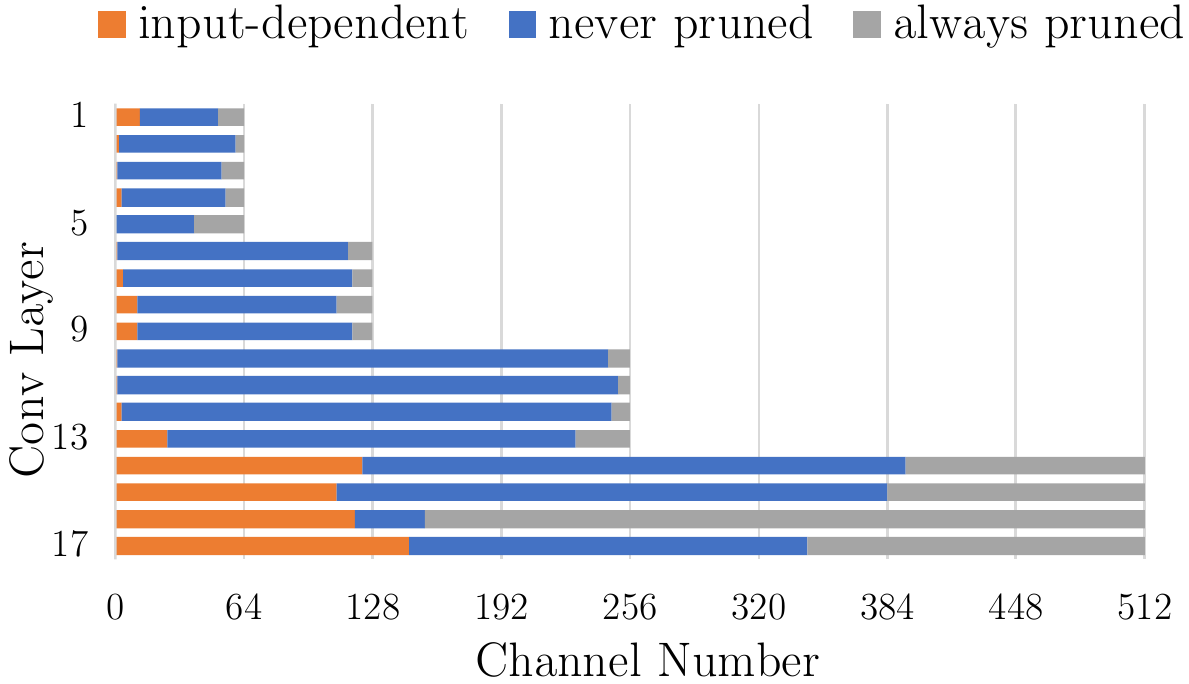}
    \caption{Distribution of channel types across different layers. The ratio of pruned channels in 16th layer is much larger than those in other layers.}
    %\caption{The distribution of channel types in each layer for ResNet-18 on \dataset{CIFAR-10}.}
    \label{fig:distribution1}
    \end{minipage}
    \hfill
    \begin{minipage}[t]{0.47\linewidth}
    \centering
    \includegraphics[height=3.4cm,width=6cm]{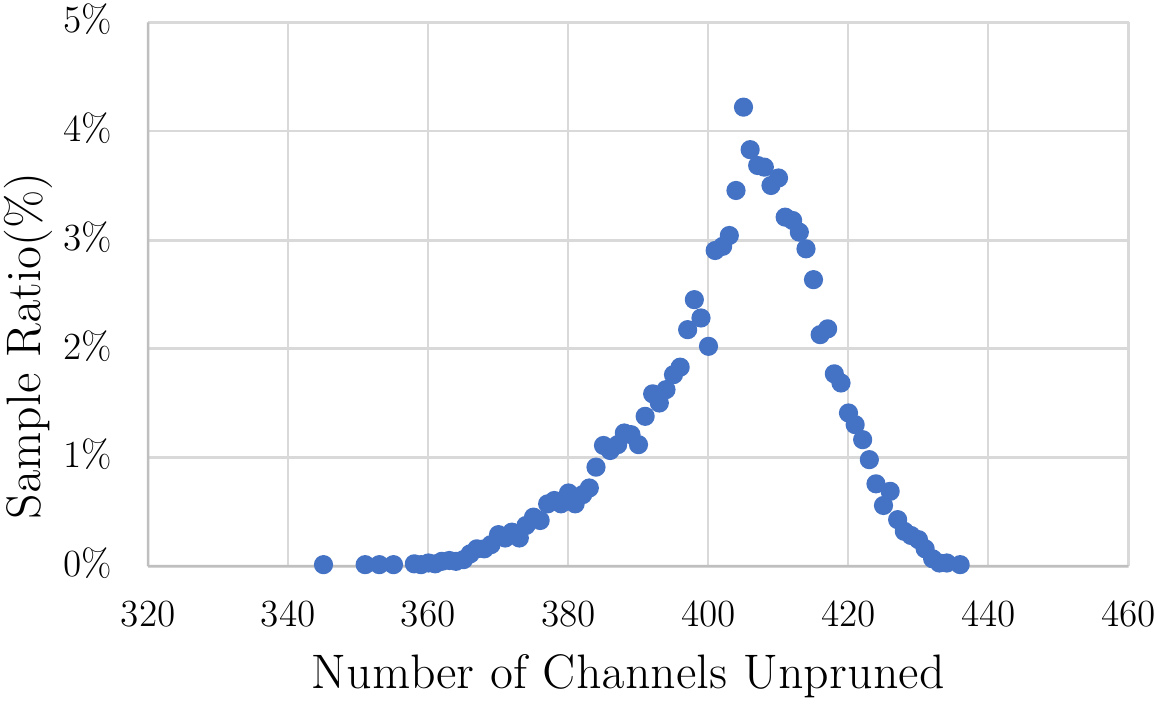}
    \caption{Distribution of unpruned channels in the last block of ResNet-18 w.r.t. samples. It is noticed that the number of unpruned channels varies much among samples.}
    %\caption{Horizontal axis is the number of unpruned channels in the last block of ResNet-18. Vertical axis is the percentage of samples in \dataset{CIFAR-10} test set.}
    \label{fig:distribution2}
    \end{minipage}
\end{figure}

Though \method outperforms its variants, it is still wondered that how the distribution of pruned channels look like. It is possible that the pruning rates are same for different layers and it is also possible that the pruned channels are same for all input samples. To answer the above questions, we investigate the pruning strategies generated by \method. We conduct forward propagation of ResNet-18 in Table \ref{tab:adaptiveness} and collect the channel pruning decisions generated by SPMs for all samples in \dataset{CIFAR-10} test set. Then we divide channels into three categories: channels that are never pruned, channels that are pruned dependent on input sample, channels that are always pruned. 

Figure \ref{fig:distribution1} shows the distribution of three types of channels. It could be clearly seen that the pruning rate varies across layers and many channels are pruned dependent on input samples, especially channels in deep layers. 
We also investigate how many channels are utilized with regard to input samples. Figure \ref{fig:distribution2} illustrates the distribution, which indicates that the channel pruning varies much based on input samples.

\section{Conclusion}
%In this paper, we propose a layer-adaptive and sample-adaptive method \method for channel pruning. We generate saliency scores and pruning strategy for channels in a layer-by-layer manner. And we propose a multi-task loss for regularizing sparsity of network to preserve accuracy and obtain expected computational cost. Experiments on \dataset{CIFAR-10} and \dataset{CIFAR-100} indicate that our model achieves state-of-the-art performance.

We propose a novel method \method for channel pruning. Our method adaptively adjusts pruning strategy for each layer according to the input samples, which enables better pruning rate and classification performance. With a differentiable design, our adaptive method can be jointly trained by classification and cost objectives, and thus maximize performance under the computational cost budget. Experiments on \dataset{CIFAR-10} and \dataset{CIFAR-100} show that our method achieves state-of-the-art performance over existing methods.

\bibliography{iclr2020_conference}
\bibliographystyle{iclr2020_conference}

\end{document}